\documentclass{article}
\usepackage{graphicx} 
\usepackage{subcaption}
\usepackage{authblk}
\usepackage{amsmath}
\usepackage{multicol}
\usepackage{caption}
\usepackage{url} 
\usepackage{makecell}
\usepackage{bbm}
\usepackage{booktabs}

\usepackage[letterpaper, total={7.5in, 9.5in}]{geometry}
\usepackage[font=small,labelfont=bf]{caption}
\newenvironment{Figure}
  {\centering\par\medskip\noindent\minipage{\linewidth}}
  {\endminipage\par\medskip}
\usepackage{xcolor}

\title{CEHR-GPT: Generating Electronic Health Records with Chronological Patient Timelines}

\author[1, 4]{Chao Pang\footnote{indicates equal contribution}} 
\author[1, 4]{Xinzhuo Jiang$^*$}
\author[1, 4]{Nishanth Parameshwar Pavinkurve}
\author[1, 4]{Krishna S. Kalluri}
\author[1, 4]{Elise L. Minto}
\author[1, 4]{Jason Patterson}
\author[2, 4]{Linying Zhang}
\author[1, 3, 4]{George Hripcsak}
\author[1, 5, 6]{Gamze Gürsoy}
\author[1, 3, 4]{Noémie Elhadad}
\author[1, 3, 4]{Karthik Natarajan}
\affil[1]{Department of Biomedical Informatics, Columbia University Irving Medical Center}
\affil[2]{Institute for Informatics, Data Science, and Biostatistics. Washington University in St Louis}
\affil[3]{Medical Informatics Services, New York-Presbyterian Hospital}
\affil[4]{Observational Health Data Sciences and Informatics}
\affil[5]{New York Genome Center, New York, NY, USA}
\affil[6]{Department of Computer Science, Columbia University}

\date{}
\begin{document}

\maketitle

\begin{multicols}{2}
\raggedcolumns

\section*{Abstract}
Synthetic Electronic Health Records (EHR) have emerged as a pivotal tool in advancing healthcare applications and machine learning models, particularly for researchers without direct access to healthcare data. Although existing methods, like rule-based approaches and generative adversarial networks (GANs), generate synthetic data that resembles real-world EHR data, these methods often use a tabular format, disregarding temporal dependencies in patient histories and limiting data replication. Recently, there has been a growing interest in leveraging Generative Pre-trained Transformers (GPT) for EHR data. This enables applications like disease progression analysis, population estimation, counterfactual reasoning, and synthetic data generation. In this work, we focus on synthetic data generation and demonstrate the capability of training a GPT model using a particular patient representation derived from CEHR-BERT, enabling us to generate patient sequences, which can be seamlessly converted to the Observational Medical Outcomes Partnership (OMOP) data format. 

\section*{Keywords}
Generative Pre-trained Transformer, Synthetic Electronic Health Records, Patient Representation, Observational Medical Outcomes Partnership - Common Data Model, Observational Health Data Sciences and Informatics

\section{Introduction}
Access to electronic health records (EHRs) is fundamental to healthcare research, drug surveillance, clinical machine learning, and system development. However, the use of real-world EHR data comes with considerable challenges such as privacy and security issues, institutional consent, and restrictions on data sharing. Synthetic data emerges as a promising solution, offering a more expedient and secure pathway to healthcare information, which could accelerate progress across various sectors, including academic research, clinical settings, and the pharmaceutical industry \cite{murray2011design}.

Synthetic data is not real data, that is, it doesn't relate to any specific individual. However, it mimics the statistical characteristics and journeys of specific patient populations. Synthetic data enables a broader range of researchers to answer their questions of interest without going through the cumbersome process of accessing real data and worrying about patient privacy \cite{Ghosheh2022}. In recent years, many machine learning, specifically deep learning and generative artificial intelligence (AI) models, have been developed to derive synthetic data from real EHR data \cite{Choi2017}. However, most existing methods for synthetic EHR data generation fail to adequately capture the temporal dependencies that are often critical in medical scenarios. These temporal aspects, such as medication schedules, symptom progression, and lab result timelines, are vital for a comprehensive understanding of patient health trajectories and for developing effective treatment strategies. An ideal synthetic dataset derived from institutional data should maintain the inherent correlations among time-series features, thus enabling researchers to externally validate machine learning models in different populations. Crucially, the synthetic dataset must preserve accurate patient timelines, as predictive tasks are highly susceptible to temporal variations. A synthetic dataset is considered to exhibit comparable machine learning utility to the original data if it meets two key criteria: 1) it demonstrates similar outcome prevalence to the source data; 2) machine learning models trained on the synthetic data achieve performance metrics akin to those trained with the original data.

The majority of the existing research focuses on developing new deep learning models in generative EHR research but without adequate emphasis on retaining accurate temporal information \cite{Li2017, Li2023, Lee2020}. Unfortunately, synthetic EHR datasets developed as such will not support use cases that require the accurate construction of a patient timeline e.g., 30-day readmission, one-year risk of heart failure, and disease progression. This limits existing work to only perform simple code prediction in their evaluations instead of comprehensive phenotype predictions. Another challenge in using synthetic EHR data in practice is its difficulty with dissemination due to a lack of standards. Synthetic patient sequences cannot be widely adopted for analyses without use of a common data model, however, none of the existing works have included such a component in their frameworks to present the synthetic data in an easy-to-consume fashion.  

In our view, time-series synthetic data should not only capture the underlying characteristics of heterogeneous EHRs but also satisfy the following temporal requirements, 1) a matching distribution of the starting age; 2) a matching distribution of the starting year; 3) a matching distribution of the inpatient duration; 4) a matching distribution of time intervals between neighboring visits. Furthermore, synthetic EHR data should be stored in common data models such as the Observational Medical Outcomes Partnership (OMOP) Common Data Model, which is used in many large data networks, \cite{Hripcsak2013} for easy dissemination and consumption. Although creating such a time-series EHR dataset seems to be a challenging task, we think that this problem can be solved through a patient representation approach. The key is to focus on designing a good patient representation rather than creating a sophisticated architecture to model time and medical events simultaneously. In this paper, we present the CEHR-GPT framework for building an end-to-end workflow to generate time-series synthetic EHR data. Our contributions are summarized below, 
\begin{itemize}
  \item We design a novel patient representation that captures visit types, discharge facilities for inpatient visits, and all temporal data, such as starting year, age, intervals between visits, and inpatient visit duration. This is the first instance of fully preserving such temporal information, to our knowledge.
  \item We treat patient sequence generation as a language modeling problem, which allowed us to use the state-of-the-art language model Generative Pre-trained Transformers (GPT) to learn the distribution of patient sequences to generate new synthetic sequences \cite{Radford, Vaswani}.  
  \item We converted synthetic sequences to the common data format OMOP with almost no loss of temporal information. Synthetic OMOP can be easily evaluated using the OHDSI tools and disseminated to others.  
  \item We evaluated the synthetic EHR data on three levels, dimension-wise distribution (marginal distribution), co-occurrence relationship, and machine learning model performance metrics. 
  \item We utilized the state-of-the-art evaluation framework proposed by Yan \textit{et al.} \cite{Yan_Brad_2022} to quantify the privacy risk by calculating a number of privacy metrics for synthetic Electronic Health Records (EHR) data, and demonstrated that the privacy risk associated with this synthetic data is low.
\end{itemize}

\section{Related work}
With the adoption of Generative Adversarial Networks (GANs) \cite{Goodfellow2014}, researchers have found creative ways to generate synthetic EHR data. Since 2017, several groups have applied GANs to tabular EHRs and developed several evaluation and privacy metrics to quantify the performance of GANs \cite{Choi2017}. Despite the success, one limitation is that the tabular format fails to capture the temporal nature of EHR data because they are constructed from patient histories using a bag-of-words approach. It was not until 2020 that researchers started developing new GAN architectures to tackle the time series data. Dual adversarial autoencoder (DAAE) \cite{Lee2020} used a combination of a variational autoencoder (VAE) and two GAN components, where the inner GAN was trained to replicate the encoded representation generated by the encoder, and the outer GAN was trained to generate realistic-looking patient sequences in addition to the reconstruction error. Another model called EHR-M-GAN \cite{Li2023} employed a similar autoencoder architecture with two main differences 1) they used a dual-VAE framework to handle the continuous and discrete valued features, where the continuous and discrete representations were generated by encoders first and then collectively used for decoding; 2) in the GAN generator, they used two parallel recurrent neural networks (one for sampling continuous noise and one for discrete noise) with a so-called Bilateral LSTM cell to allow the continuous and discrete noise vectors to interact with each other to generate better sampling vectors. Although these GANs took the temporal order of events into consideration, they did not generate timestamps for visits or medical events, therefore limiting the use of the synthetic data. 

Improving upon the previous works, a two-stage learning algorithm (dependency learning and conditional simulation) named SynTEG was proposed to generate timestamped synthetic data \cite{Zhang2021}. In dependency learning, transformer encoders learned visit representations, which were used to feed into an recurrent neural network (RNN) model to learn the dependencies between visits. Two self-supervising learning tasks were used for training – prediction of the timestamp and diagnoses at the next visit. In the conditional simulation, the learned visit representations generated from the previous step were used to train a conditional GAN to generate the diagnosis codes at each visit. This approach achieved superior performance in the time-sensitive evaluations over the previous methods. However, there are still ongoing challenges that have not been addressed in their work, 1) other EHR data was not utilized because only diagnosis codes were included in training; 2) the visits were assumed to start and end on the same day, which would fail to model in-patient visits that normally span days; therefore, the constructed patient timeline would be inaccurate; 3) the synthetic data didn't include visit types and discharge facilities. 

Until now, almost all the existing approaches used some variation of GAN for learning the data distribution, but unfortunately, GANs are notoriously difficult to train and easily subject to mode collapse. Despite the recent advancements in optimization techniques such as Wasserstein-GAN \cite{Gulrajani2017}, it would require a significant amount of time to tune hyperparameters to train GANs. As the previous works have demonstrated \cite{Pang2021, Li2020, Rasmy2021}, patient sequence generation could be conceptually represented as a language modeling problem. Foresight \cite{Kraljevic2022} adapted GPT to forecast patient trajectories. Their method used a name entity recognition (NER) tool to extract medical concepts from discharge summaries, based on which a patient sequence was constructed chronologically. Then they trained a standard GPT model using all patient sequences constructed from the previous step. To forecast future events, they fed a patient history to prompt the model and employed a Monte Carlo sampling strategy to calculate the probability of developing certain conditions. One limitation of this work was that the model could not predict when a certain condition would happen due to a lack of temporal information in their patient sequence. Nevertheless, their work demonstrated the potential of using GPT for modeling patient sequences.  We aim to address these limitations in our work. 

\vspace{20pt}

\begin{Figure}
    \includegraphics[width=\linewidth]{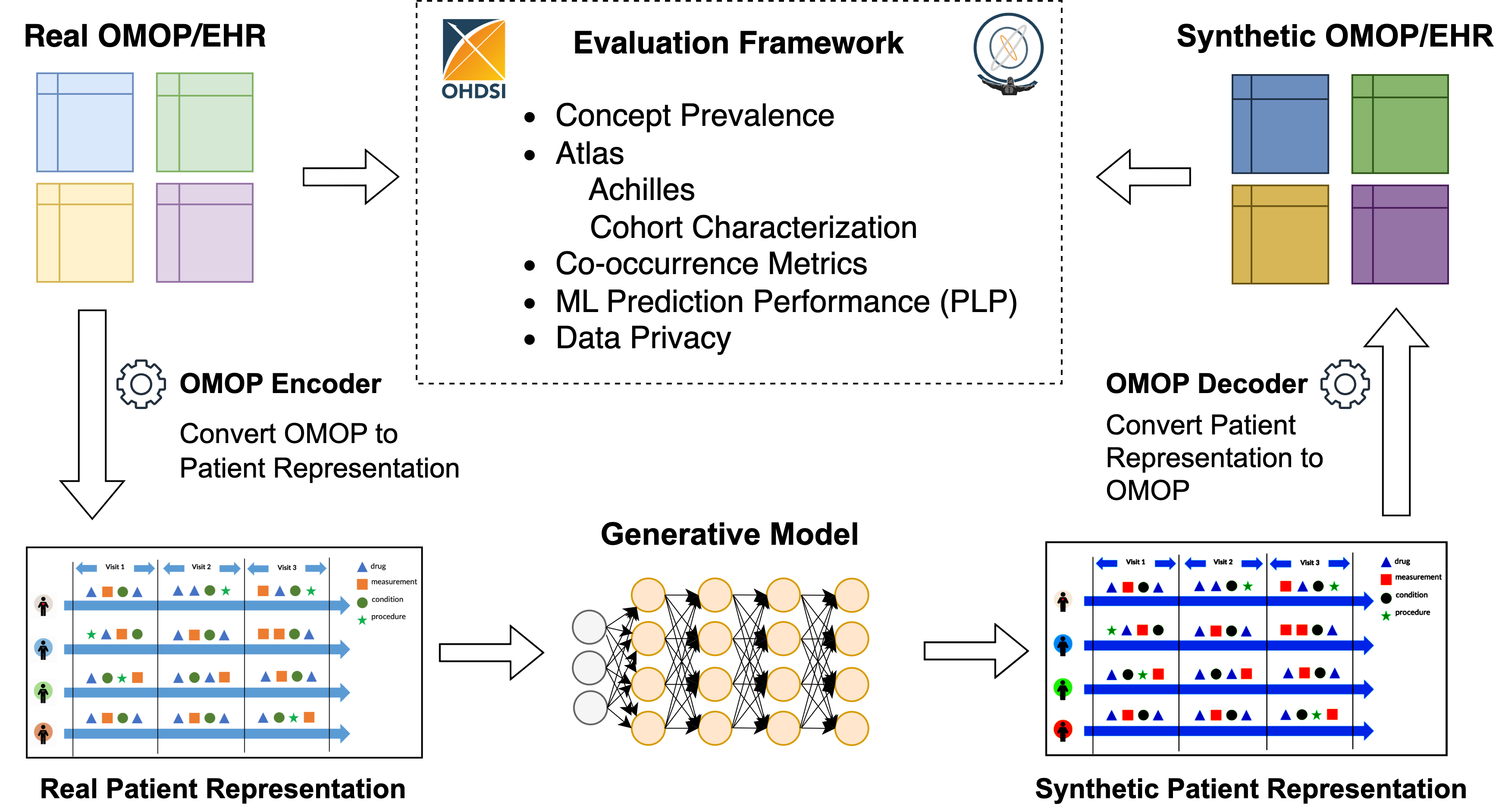}
    \captionof{figure}{The OMOP data is first converted to patient sequences by an OMOP encoder based on the patient representation that preserves demographics, visit types, and temporal intervals between visits. Then a generative model is trained to learn the sequence distribution to generate new sequences. Next, the generated sequences are converted back to the OMOP format using an OMOP decoder.}
    \label{figure1}
\end{Figure}

\section{Methods}
In Figure \ref{figure1}, we present the framework for generating synthetic EHR data from an OMOP source. To retain the temporal dependencies, we opted to work directly with time-series patient sequences instead of using a bag-of-words (BOW) representation by building on our previous work \cite{Pang2021}. We first encoded the OMOP data into patient sequences using a specific patient representation, described later in this section. Secondly, we trained a generative model on the converted patient sequences and utilized it to generate new synthetic patient sequences. Finally, we fed the synthetic patient sequences into an OMOP decoder to create a synthetic OMOP dataset. Furthermore, an evaluation procedure was developed to assess the similarity between the synthetic OMOP and the source OMOP data.

\subsection{Patient Representation}
\begin{figure*}
    \centering
    \includegraphics[width=0.9\linewidth]{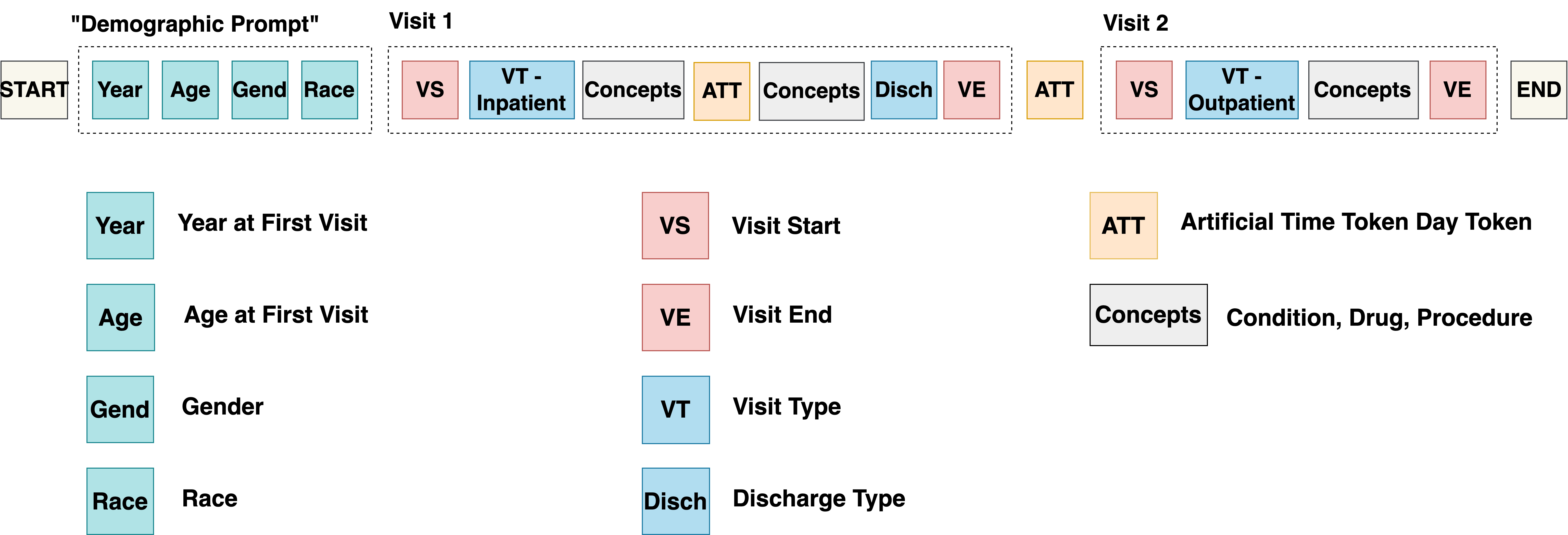}
    \captionof{figure}{The patient representation preserves demographics, visit types, and temporal intervals between visits and inpatient duration. It's designed to have the demographic prompt at the beginning including year at the first visit, age at the first visit, gender and race tokens, then followed by a series of visit blocks to represent the complete patient timeline. An artificial time token (ATT) is inserted between the neighboring visit blocks to keep track of the time intervals in days. In each visit block, all the essential information is retained including the visit type and domain records. In the case of inpatient visits, the inpatient ATT tokens (representing time intervals in days) are inserted between groups of concepts that occur on the same day, in addition, a discharge token is provided at the end of the visit block. }
    \label{patient_representation}
\end{figure*}

We designed a patient representation in CEHR-BERT \cite{Pang2021} that captures medically relevant events and their timelines while exhibiting certain characteristics of a sentence. In order to fully leverage Large Language Models (LLM) on patient sequences, we further extended this patient representation to include demographic information, patient history, and temporal dependencies as shown in Figure \ref{patient_representation}. The start of the sequence defines the demographic prompt containing EHR start year, age, gender, and race. It is followed by visit blocks separated by artificial time tokens (ATT) representing time intervals in days, e.g., \(D_1\) represents an interval of 1 day.  For time intervals surpassing 1080 days, we grouped these into a single Long Term (LT) token, a decision guided by the low occurrence rate in this time frame. Each visit block starts with a visit type token (VTT) to signify the type of visit and is then followed by domain records arranged in chronological order. In the case of inpatient visits, a distinct inpatient ATT (IATT) was inserted between neighboring inpatient spans, defined as the groups of records that occurred on the same day. These are distinct from ATT and are used for capturing the time between multiple events characterized by concepts within the same visit. In addition, a discharge facility code (e.g. discharge home and long-term care) was inserted at the end of the inpatient visit. The use of IATT, which is distinct from ATT, was necessary since they are attributed to two different contexts, and resulted in better performance than using the same ATT across both contexts.

This patient representation allows us to convert from any common data model (e.g. OMOP) to patient sequences and vice versa without any loss of temporal information. To formulate this property, let's denote \(D_i\) to be data associated with the \(i\)th patient in the source format, \(P_i\) to be the patient sequence converted from \(D_i\), and \(F\) to be the function that converts source data to patient sequences, represented as \(P_i = F(D_i)\). Let's then denote \(D_i'\) to be the reconstructed source data and \(F'\) to be the inverse function that converts patient sequences back to original source format. This indicates \(D_i' = F'(P_i)\). Finally, let's denote \(T\) to be a function that extracts all the dates for a set of patient records. The patient representation is said to preserve temporal information perfectly if and only if the following statement is true for every single patient, \(T(D_i) = T(D_i') + C_i\), where \(C_i\) is a constant that represents a consistent time shift e.g., \(C_i=4\) days.

\subsection{OMOP Encoder}
To create a patient sequence, we began by generating a demographic prompt using data from the OMOP person and visit tables. This prompt included essential demographic information such as the patient's age at their initial visit, the year of their first visit, their gender, and their race. Subsequently, we constructed a series of visit blocks to represent the patient's entire medical history. We inserted an ATT token between these visit blocks to signify the time intervals between them. Within each visit block, we gathered all relevant records from OMOP domain tables (e.g., condition) and arranged them chronologically based on their respective timestamps. In cases where there are timestamp ties, we sorted the concepts. Additionally, three artificial tokens (VS, VE, and VTT) were added at the beginning and end of each visit block to denote the start, end, and type of the visit, as illustrated in Figure \ref{patient_representation}. For inpatient visit blocks, extra processing steps were necessary. Initially, we grouped records by their timestamps to identify all inpatient spans, arranging them in chronological order. Next, we inserted IATT between these spans, aligning them with the respective time intervals. Finally, the discharge facility code was extracted from the OMOP visit table and appended to the end of the block.

\subsection{Generative Model}
We used a GPT model with standard transformer decoders, where the input layer utilized concept embedding and trainable positional embedding. The model was trained using the Next Word Prediction learning objective. When generating a patient sequence, we randomly sampled a demographic prompt from our source sequences, which served as the input to the GPT model. Using these prompts, the entire patient history was generated autoregressively by sampling tokens from the predictive distribution at the final layer. 

\subsection{OMOP Decoder}
The patient sequence was converted back to the OMOP format using the OMOP decoder. The start-year prompt determined the EHR history's beginning, using January 1st as the default. Demographic data was stored in the person table, while concepts were transformed into condition, drug, and procedure tables. A date cursor was used to represent the “current time” as we were processing each patient sequence, it was initially set to the star year and was updated whenever an ATT token was encountered. We first parsed out the number of days represented by the ATT token, and then moved the data cursor by the same number of days to the future. During the sequence processing, the VS token marked the start of a new visit block. We, therefore, extracted the token corresponding to the visit type (the token immediately followed by VS) and created a new visit record with the corresponding type. 

All the tokens subsumed by this visit block were converted to condition, drug, and procedure records and linked to the current visit. For outpatient visits, we assumed that all data points were generated on the same day; therefore, we set the start and end dates of the visit to the current value of the date cursor. Similarly, all the domain records were set to the date cursor as well. 

When processing inpatient visits, the date cursor was updated inside the visit block based on the time intervals represented by IATT tokens between inpatient spans. Domain records were generated with the current value of the date cursor. Towards the end of the inpatient visit block, we extracted the last token (the token right before VE) corresponding to a discharge facility and updated the visit end date using the current value of the date cursor (as the date cursor was frequently updated inside an inpatient visit block). This allowed us to preserve the complete information about inpatient visits. If generated sequences do not follow patterns presented in the patient representation, they will be discarded to ensure the quality of the synthetic data.

\section{Experiments and Results}
\subsection{Data and Preprocessing} \label{data_processing}
The source patient sequences were generated from the OMOP database derived from Columbia University Irving Medical Center-New York Presbyterian Hospital EHR data, which includes 3.7 million unique patients' medical histories including condition, medication, and procedure. Unknown concepts (i.e., \(concept\_id=0\)) were removed from all domains except for the visit type when constructing the patient sequences using the proposed patient representation. Patients with less than 20 tokens were removed from the training dataset, and approximately 2.3 million patients were included for training whereas 75,000 patients were held out for privacy evaluations. For the GPT model, we used a context window of 512, 16 transformer decoders, 8 attention heads with a dropout rate of 0.1, and 128 dimensions for both the embedding and hidden units. All patients with longer than 512 tokens were post-truncated to fit the context window. The statistics of training data was summarized in Table \ref{tab:summary-stats}. We trained the model for 2 epochs on 2 Nvidia 2080 TI GPUs with a batch size of 32 and a learning rate of 0.0002. The model checkpoint was created every 10,000 steps. 

\begin{center}
\captionsetup{justification=raggedright,singlelinecheck=false}  
\renewcommand{\arraystretch}{1.2} 
  \begin{tabular}{cp{3.5cm}p{3.5cm}}
  \toprule 
  \textit{}  & 
  \textit{No. of visits per patient} & \textit{Sequence length per patient}\\
  \midrule
  mean & 16 & 148\\
  std & 19 & 154\\
  min & 2 & 20\\
  25\% & 4 & 38\\
  50\% & 8 & 78\\
  75\% & 21 & 198\\
  max & 102 & 512\\
    \hline
  \end{tabular}
  \captionof{table}{Summary statistics of the CUIMC-NYP OMOP training data}
  \label{tab:summary-stats}
\end{center}

During the first epoch, we used a standard data generator strategy, where every training example was fed to the model; however, we switched to a random sampling strategy to draw training examples during the second epoch. For synthetic data generation, we used the 10th model snapshot because early experiments showed its superior performance compared to other snapshots. Furthermore, we used several sampling hyper-parameters including top k=300, top k=200, top k=100, top p=90\%, top=95\%, and top=100\% to generate different synthetic OMOP datasets. Using specific top p/k values is a common technique in language models for data generation. For instance, the top k approach limits the selection of the k most probable tokens in the prediction distribution during sampling. On the other hand, the top p method selects a set of most likely tokens whose combined probability reaches p\% (e.g. 90\%) in the predictive distribution. For each sampling strategy, 1M synthetic sequences were generated and converted to OMOP. On average, 98\% of the generated sequences passed the validation and were converted to OMOP.

\subsection{Evaluations}
We followed the evaluation procedures proposed by \cite{Yan_Brad_2022} to compute the data utility metrics including dimension-wise distribution, co-occurrence relationship, and machine learning model performance. Some of the metrics were originally designed for tabular EHR data; therefore, we adapted them to the time-series setting. When using Kullback-Leibler (KL) divergence to evaluate source and synthetic datasets, we use concept probabilities defined as,
\[
P_{prob}(c)=\frac{\sum_i^n \mathbbm{1}\Big[c \in h_i \Big]}{\sum_i^n \sum_j^m \mathbbm{1} \Big[c_j \in h_i \Big]}
\]
where $c$ denotes the target concept, $h_i$ denotes the $ith$ patient history, $n$ and $m$  denote the total number of patients and concepts respectively. Due to the small probability values, we opted to use the prevalence instead for data visualization using a slightly modified formula below, 
\[
P_{prev}(c)=\frac{\sum_i^n \mathbbm{1}\Big[c \in h_i \Big]}{n}
\]
The only difference between the prevalence and the probability is the normalization constant used as the denominator. 

For baseline comparison, we added three variants of GPT models trained on slightly different patient representations. The first baseline model was trained on the adjusted CEHR-BERT representation, which differs from the proposed patient representation by 1) CEHR-BERT ATT tokens contain a mix of day/week/month/year tokens to represent time intervals while CEHR-GPT only used the day tokens and 2) the IATT tokens and discharge facility tokens were not used in the CEHR-BERT representation. The second baseline model used the proposed patient representation with IATT tokens removed. We will refer to this baseline as GPT-OUTPAT. The last baseline GPT was trained on the patient representation widely used in time-series EHR research, where the concepts were simply ordered chronologically and put in a sequence without any additional artificial tokens. To use such sequences for comparison, we assumed that all medical events in the patient sequence belonged to a single visit. This baseline model will be referred to as GPT-Vanilla. We only used the top\_p=95\% sampling strategy to generate synthetic OMOPs for these baseline models. The comparison between these patient representations can be seen in Supplementary Figure \ref{patient_history_comparison}.

\subsubsection{Dimension-wise Distribution}
KL divergence was assessed to compare the concept probability distributions between synthetic and source datasets among the entire population. In Figure \ref{concept_kl_divergence}, synthetic datasets generated by different patient representations and sampling strategies were evaluated against real patient data. The results showed that baseline models CEHR-BERT, GPT-Vanilla and GPT-OUTPAT with sampling strategy using the threshold of top\_p$=95\%$ diverged the least from real concept probability distributions followed by GPT models trained with top\_p$=100\%$ and top\_p$=95\%$. GPT models with threshold of top\_k$=300$ and top\_k$=200$ with relatively similar divergence. However, GPT models with threshold of top\_k$=100$ sampling strategy had the largest KL divergence.

\begin{Figure}
    \includegraphics[width=\linewidth]
    {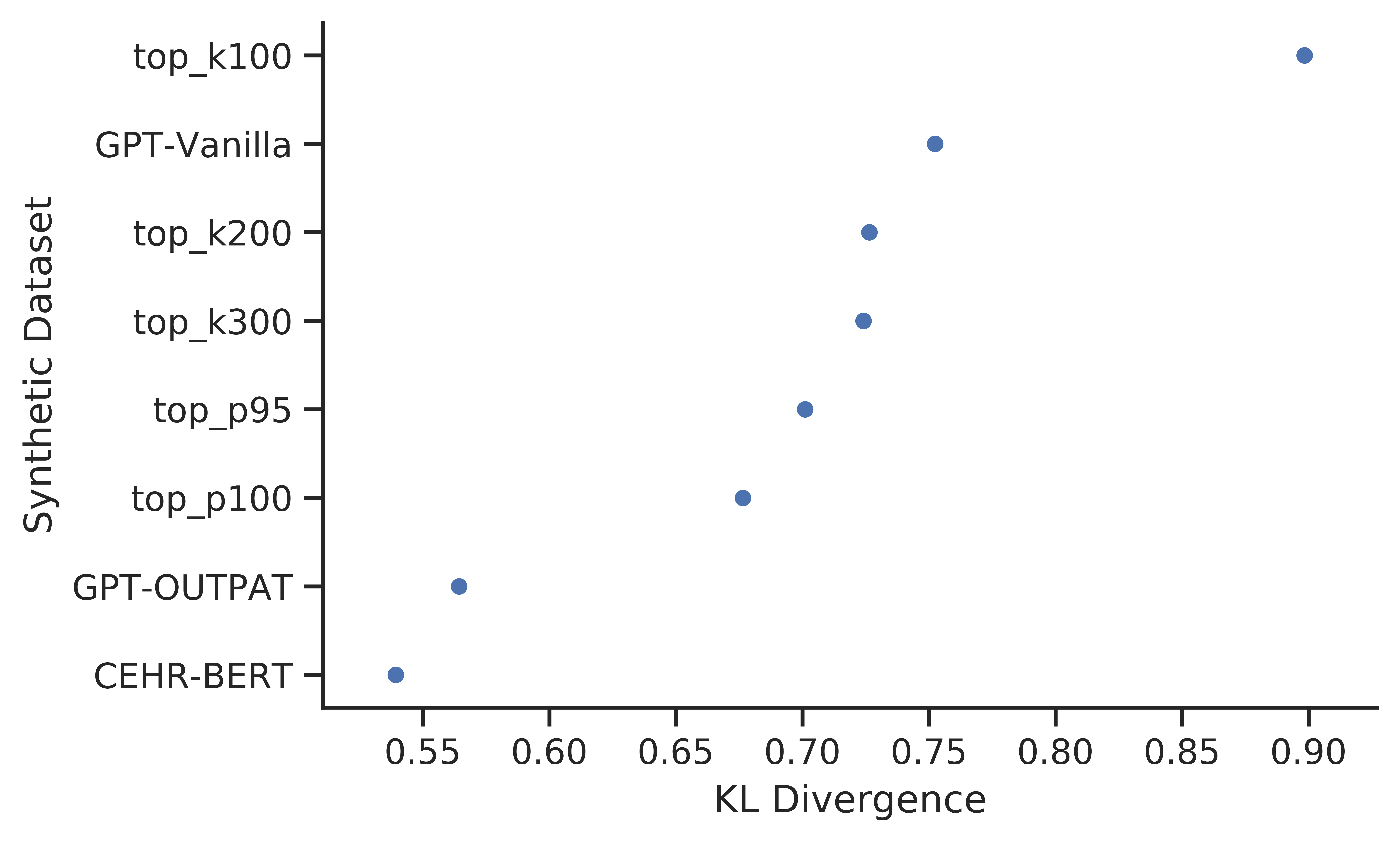}
    \captionof{figure}{KL divergence for comparing concept probability distribution between synthetic data and real data. The probabilities of concepts were calculated on the scale of the entire population.}
    \label{concept_kl_divergence}
\end{Figure}

Furthermore, we conducted a qualitative analysis using the synthetic data top\_p=95\% to gain a more comprehensive understanding. The dimension-wise distributions between the synthetic and source datasets were compared at three distinct levels: the entire population, specific sub-groups (e.g., female population), and particular cohorts (e.g., hospitalization cases). Figure \ref{concept_prevalence} illustrates the concept prevalence comparison between the original OMOP dataset and the generated OMOP dataset, using a threshold of top\_p$=95\%$. In the high-frequency regions, most data points cluster closely around the diagonal line, indicating a strong agreement between the source and synthetic data. Conversely, data points appear more dispersed for low-frequency concepts. Notably, in the subplot representing female conditions (located in the first column, second row), there is an unusual cluster of concepts positioned above the diagonal line. Further examination revealed that these concepts were male-specific and should not appear in the female population. Although there are a few instances of such cases in the source data, GPT amplified such cases in synthetic data.

\begin{Figure}
    \includegraphics[width=\linewidth]{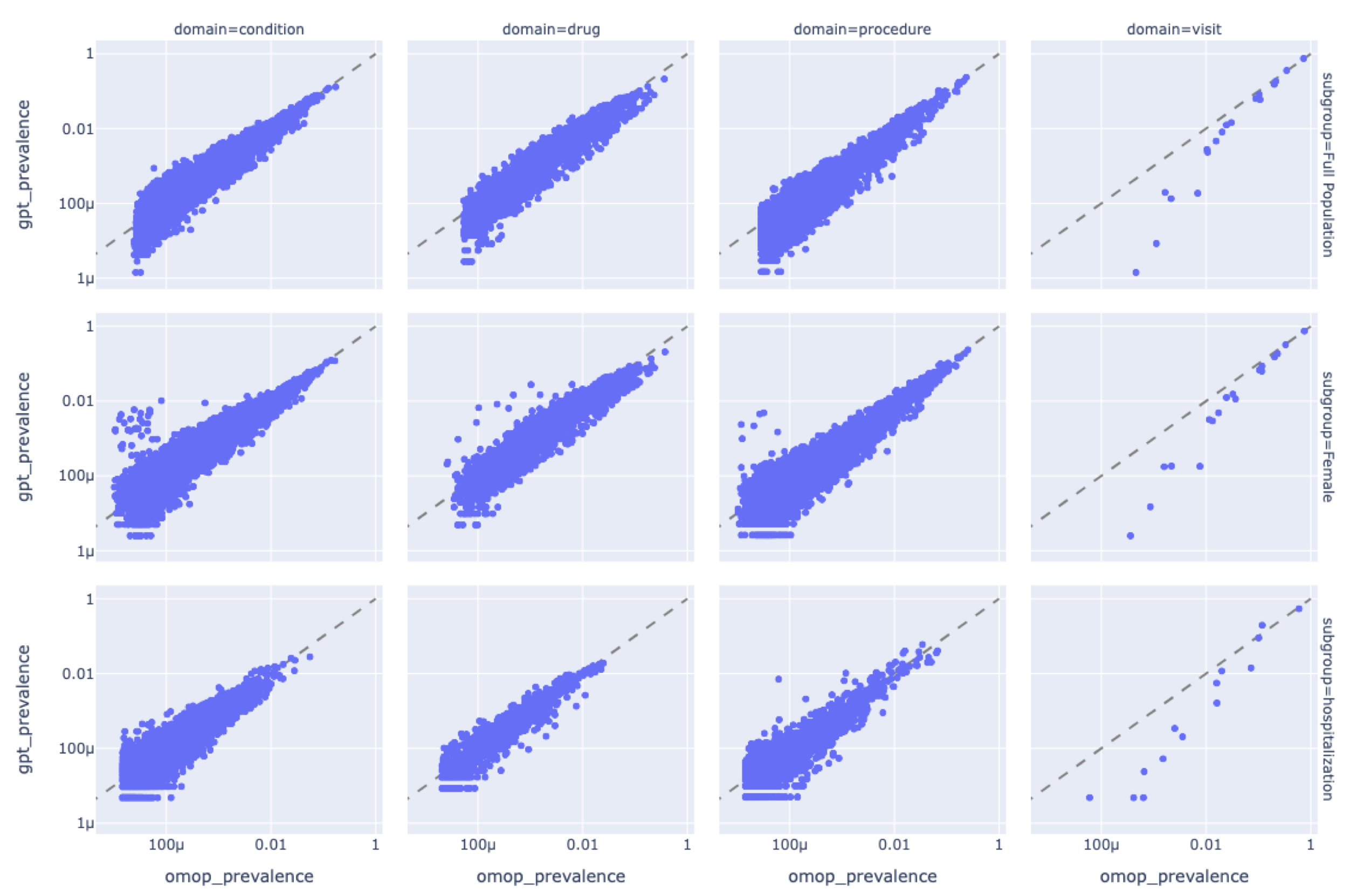}
    \captionof{figure}{Concept prevalence comparison between the source OMOP and generated OMOP using top $p=95\%$ in the log scale stratified by domain in columns and by population in rows, where x-axis and y-axis represent the source and the synthetic data respectively, and each dot represents a concept}
    \label{concept_prevalence}
\end{Figure}

In addition, we conducted a detailed comparison of the visit tables between the two OMOP datasets. Our specific focus was on performing demographic breakdowns and analyses for gender, race, and age group. The supplementary Figures \ref{visit_age_breakdown}, \ref{visit_gender_breakdown}, \ref{visit_race_breakdown}  highlight the top 10 most prevalent visits for each demographic breakdown, showcasing that the trends in both the source and generated visit tables exhibited notable similarities.

\subsubsection{Co-occurrence Relationship}
To measure how closely the generated datasets resemble the source, we computed the KL divergence between their co-occurrence matrices. The matrix was constructed temporally with the following logic: 1) for each concept in the patient sequence, only the future concepts were taken into consideration for creating concept pairs; 2) each patient could only contribute to the same concept pair once; 3) the matrix was normalized into a proper probability distribution by dividing the occurrences of each concept pair by the overall number of pairs. Additionally, we set benchmarks for our analysis: a lower-bound and an upper-bound. The lower bound was determined by applying the KL divergence method to two random samples from the source data. The upper bound was established by creating a theoretical co-occurrence matrix under the assumption that all concepts in the source data were independent. The KL divergence was then applied to this hypothetical matrix to calculate the upper bound. Figure \ref{cooccurrence_cleveland} illustrates that the datasets with CEHR-BERT and top\_k=300 most closely approach the lower bound, with the top\_p=95\% and GPT-OUTPAT baselines coming next. Datasets with top\_k=100 and top\_k=200 had marginally higher KL divergence values, while the top\_p=100\%, the GPT-Vanilla datasets exhibited the largest KL divergence. 

To thoroughly examine the similarities between the original and the synthetic data, we carried out a qualitative analysis using one of the synthetic datasets. This involved comparing the most frequent pairs of co-occurring concepts within each category, such as condition-condition (interpreted as one condition concept followed by another condition).

\begin{Figure}
    \includegraphics[width=\linewidth]{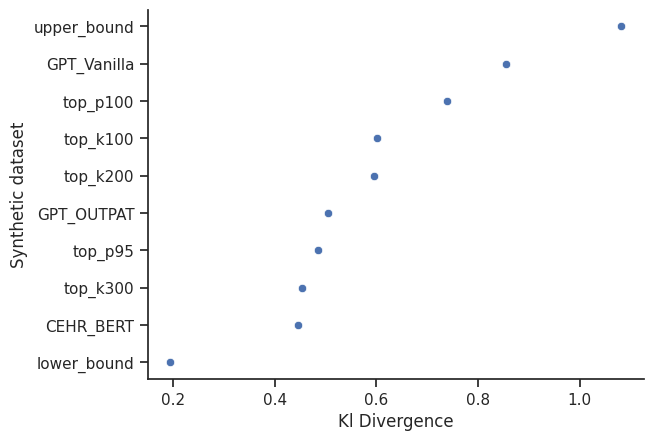}
    \captionof{figure}{KL divergence associated with different synthetic data. The closer to the lower bound in the bottom left corner, the better the synthetic data.}
    \label{cooccurrence_cleveland}
\end{Figure}

\begin{Figure}
    \includegraphics[width=\linewidth]{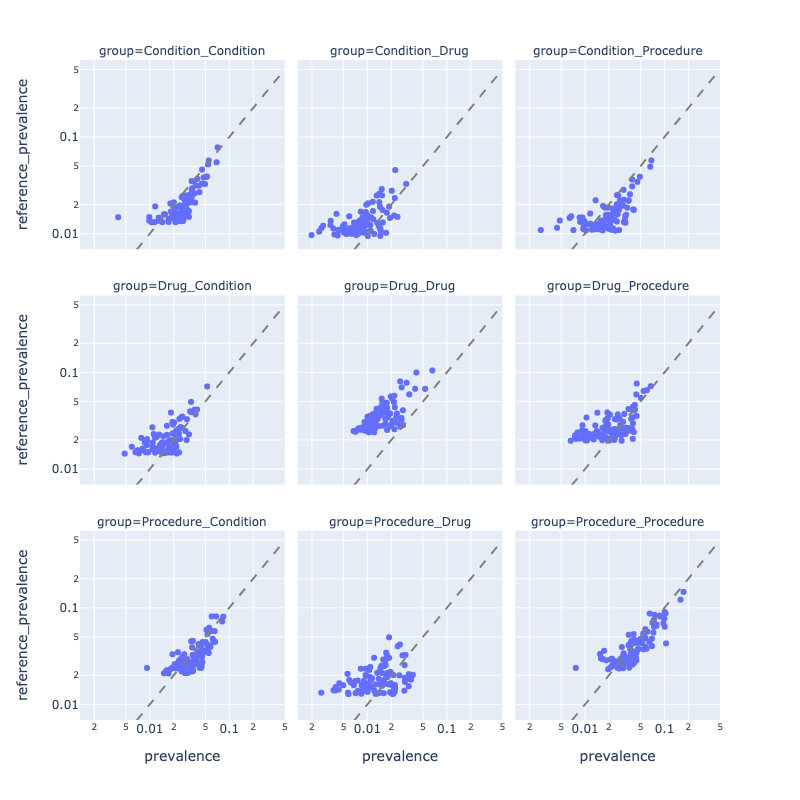}
    \captionof{figure}{Top 100 co-occurring concept pairs for each co-occurrence category e.g. condition-condition interpreted as a condition concept followed by another condition concept. The x and y axes represent the synthetic and source data respectively}
    \label{concept_conditional_prevalence}
\end{Figure}

The results, illustrated in Figure \ref{concept_conditional_prevalence}, display the top 100 pairs for each type of co-occurrence relationship. The analysis revealed that the synthetic data accurately mirrored the co-occurrence patterns in most categories, with the exception of the categories ending with a drug concept (as shown in the second column of Figure \ref{concept_conditional_prevalence}), where the data points were more scattered. Notably, a majority of the points in condition-condition and procedure-procedure pairs aligned along the diagonal line.

\begin{Figure}
    \includegraphics[width=\linewidth]{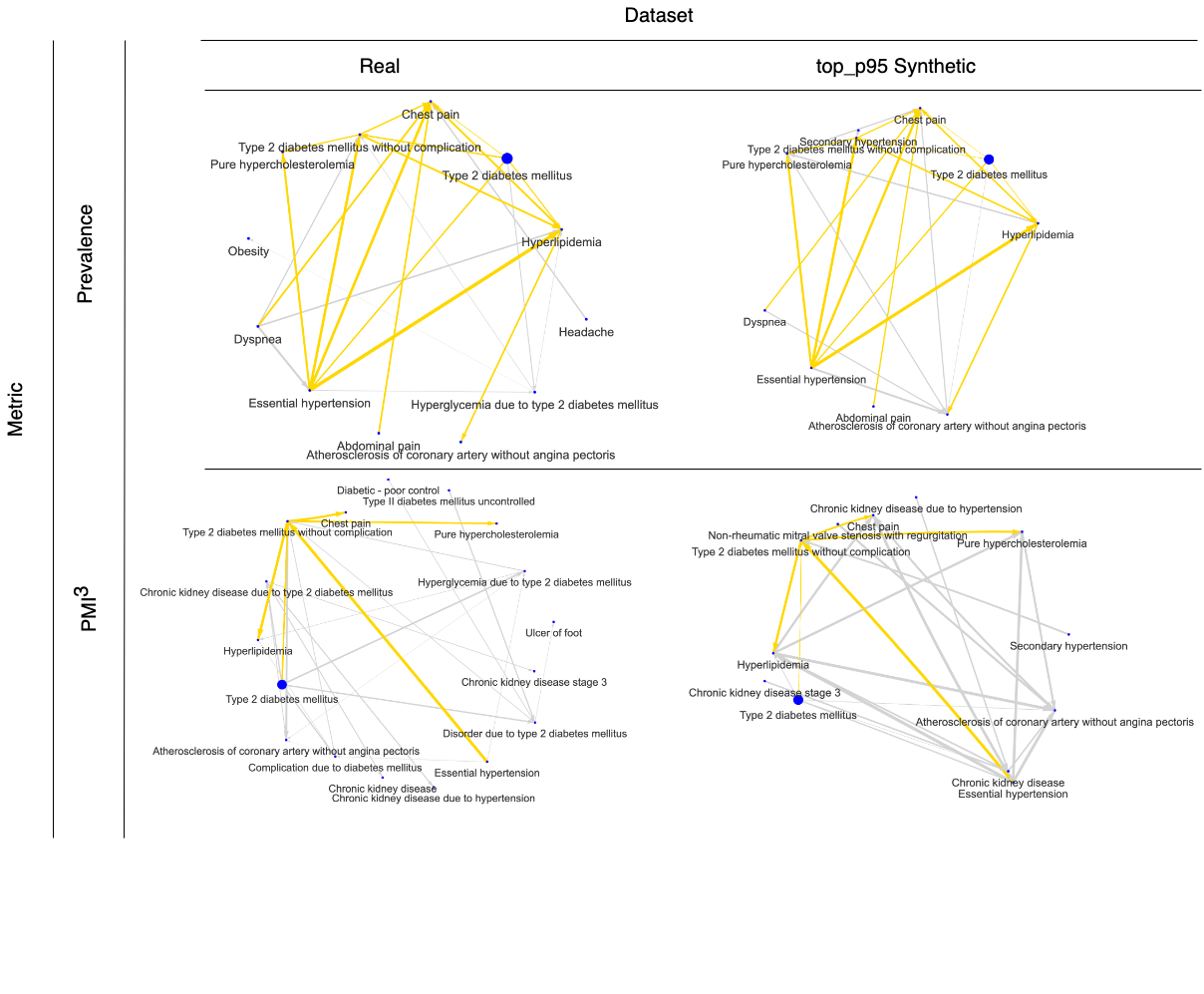}
    \captionof{figure}{Comparison of temporal co-occurrence networks around Type 2 Diabetes Mellitus (T2DM) for real (left) and synthetic data (right). Networks were constructed with T2DM's (large circle) top 5 co-occurring condition concepts and their own top 5 co-occurring condition concepts. The metrics Prevalence (top) and Pointwise Mutual Information-3 (PMI$^3$) (bottom) were used to quantify co-occurrence. Gold edges indicate edges that are shared by both real and synthetic data, edge thickness indicates the strength of co-occurrence, and arrow direction indicates the direction of the temporal association.} 
    \label{Diabetes_network}
\end{Figure}

For an in-depth analysis, we focused on examining the co-occurrence relationships associated with Type 2 Diabetes Mellitus (T2DM). Figure \ref{Diabetes_network} qualitatively compares the real and top\_p=95\% synthetic data networks of condition concepts that co-occur around T2DM. The degree of co-occurrence is calculated with prevalence, which evaluates the frequency of concept co-occurrence, and Pointwise Mutual Information-3 (PMI$^3$), which evaluates the probabilistic association of concepts. There is much overlap between the real and synthetic networks when prevalence is used to construct them, but connections are more disparate when PMI$^3$ is used. 

\subsubsection{Predictive performance}
For this analysis, we constructed five prediction tasks using the method described in \cite{Pang2021} and the Book of OHDSI \cite{ohdsi2019book}. Table \ref{tab:cohort_definition} shows the cohort and the corresponding definition.
\begingroup
\setlength{\tabcolsep}{10pt} 
\renewcommand{\arraystretch}{1.8} 
\begin{table*}[t]
  \centering
  \begin{tabular}{p{4.0cm}p{12.0cm}}
  \toprule 
  \textit{Cohort}  & \textit{Definition}\\
  \midrule
    HF readmission & HF patients who have a 30-day all-cause readmission. Observation window: 360 days, Prediction windows 30 days \\
    Hospitalization & 2-year risk of hospitalization starting from the 3rd year since the initial entry into the EHR system. Observation window: 540 days, hold-off window: 180 days, prediction windows: 720 days  \\
    COPD readmission & COPD  patients who have a 30-day all-cause readmission. Observation window: 360 days, prediction windows: 30 days \\
    Afib ischemic stroke & Afib patients with 1-year risk since the initial diagnosis of afib ischemic stroke. Observation window: 720 days, prediction windows: 360 days \\
    CAD CABG & Patients initially diagnosed with Coronary Arterial Disease (CAD) without any prior stent graft will receive the  Coronary artery bypass surgery (CABG) treatment. Observation window: 720 days, prediction windows: 360 days\\
    \hline
  \end{tabular}
  \caption{Cohort definitions}
  \label{tab:cohort_definition}
\end{table*}
\endgroup
To extract features, we first rolled up the medical concepts using ontological hierarchies to reduce dimensionality [see supplementary materials] and used the bag-of-word (BOW) approach, where we counted the frequency of each concept in a given observation window. For each task, we split the cohort data (both synthetic and real) into training and testing sets with a split ratio of 85:15. We ran logistic regression using Sklearn's implementation with the default configuration. Finally, the area under the receiver operating characteristics curve (AUC) was calculated using the test set. In addition, we reported PR-AUC (precision-recall) due to the class imbalance often present in EHR data. Table \ref{tab:prediction_model_results} shows the prevalence of the positive cases, ROC-AUC, and PR-AUC for each synthetic data. The metrics associated with the baseline CEHR-BERT are reported in the Supplementary Materials Table \ref{tab:supple_prediction_model_results}. For easy comparison of synthetic datasets, we defined a consolidated distance metric as the weighted average of the relative differences of the three aforementioned metrics,
\[
    dist = \frac{|\delta_{Pre}|\times 0.5}{Pre_{true}} + \frac{|\delta_{AUC}| \times 0.25}{AUC_{true}} + \frac{|\delta_{PR}| \times 0.25}{PR_{true}}
\]
where $\delta_{Pre}$, $\delta_{AUC}$, and $\delta_{PR}$ represent the differences in prevalence, ROC-AUC, and PR-AUC between the source and the synthetic data; $Pre_{true}$, $AUC_{true}$, and $PR_{true}$ denote the ground truth metrics generated from the source data.

Figure \ref{ml_combined_metrics} presents the distance metrics for various synthetic datasets across different cohorts. The dataset created with top\_k=300 displayed the best performance in HF and COPD Readmission but was less effective in other cohorts. The top\_p=95\% dataset maintained consistent performance levels, ranking as the best in Hospitalization and Afib ischemic stroke,  and as second-best in HF and COPD readmission. In comparison, the GPT-OUTPAT baseline dataset exhibited a slightly higher divergence across the Hospitalization, Afib Ischemic Stroke, and CAD CABG groups. However, the CEHR-BERT baseline showed similar patterns in most cohorts, with a notably lower divergence in CAD CABG. It is noted that both GPT-OUTPAT and CEHR-BERT were not included in the HF Readmission and COPD Readmission analyses due to the absence of synthetic patients meeting the selection criteria for these cohorts. Other synthetic datasets, specifically those generated with top\_p=100\% and top\_k=200 lagged behind in performance. Interestingly, top\_k=100 showed a unique pattern, being the closest in the distance for the CAD CABG cohort but underperforming in the others.

\begin{Figure}
    \includegraphics[width=\linewidth]{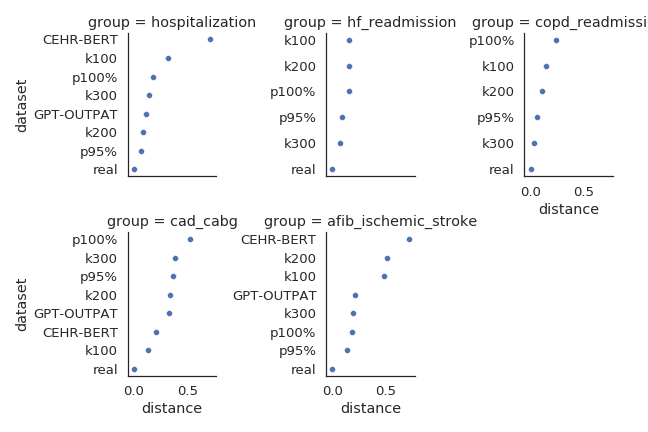}
    \captionof{figure}{The consolidated distance metrics for different synthetic datasets stratified by cohort. GPT-Vanilla and CEHR-BERT were omitted as the cohorts couldn't be constructed due to the loss of temporal information. GPT-OUTPAT was omitted from HF readmission and COPD readmission as the cohorts for the same reason.}
    \label{ml_combined_metrics}
\end{Figure}

\begingroup
\setlength{\tabcolsep}{4pt} 
\footnotesize
\begin{table*}[hbt]
  \centering
  \begin{tabular}{c|c|cccccc}
  \toprule 
  \textit{Cohort}  & \textit{Real} & \textit{p=95\%} & \textit{p=100\%} & \textit{k=100} & \textit{k=200} & \textit{k=300} & \textit{GPT-OUTPAT} \\
  \midrule
HF readmission & \makecell{\\Pre = 25.7\\AUC = 65.7\\PR = 39.3} & \makecell{\\Pre = 27.6\\AUC = 69.2\\PR = 45.7} & \makecell{\\Pre =27.7 \\ AUC = 52.4 \\ PR = 29.0} & \makecell{\\Pre = 30.7 \\ AUC = 68.1 \\ PR = 47.8} & \makecell{\\Pre = 29.3 \\AUC = 54.0 \\PR = 32.9} & \makecell{\\Pre = 26.5 \\ AUC = 61.1 \\ PR = 33.8} & \makecell{\\Pre =100.0 \\ AUC = NA \\ PR = NA}\\
    Hospitalization & \makecell{\\Pre = 5.6 \\ AUC = 75.3 \\ PR = 19.5} & \makecell{\\Pre = 5.2 \\ AUC = 77.1 \\ PR = 21.4} & \makecell{\\Pre = 7.4 \\ AUC = 71.3 \\ PR = 20.2} & 
    \makecell{\\Pre = 2.8 \\ AUC = 87.0 \\ PR = 22.1} &
    \makecell{\\Pre = 5.2 \\ AUC = 84.2 \\ PR = 20.8} & \makecell{\\Pre = 6.3 \\ AUC = 78.7 \\ PR = 24.6} & \makecell{\\Pre = 5.2 \\ AUC = 70.2 \\ PR = 14.3}\\
    COPD readmission &  \makecell{\\Pre = 34.5 \\ AUC = 74.2 \\ PR = 83.8} & \makecell{\\Pre = 37.8 \\ AUC = 76.4 \\ PR = 84.4} & \makecell{\\Pre = 47.2 \\ AUC = 74.1 \\ PR = 67.2} & 
    \makecell{\\Pre = 26.4 \\ AUC = 75.9 \\ PR = 90.3} &
    \makecell{\\Pre = 28.3 \\ AUC = 70.1 \\ PR = 82.8} & \makecell{\\Pre = 34.5 \\ AUC = 68.8 \\ PR = 80.2} & \makecell{\\Pre = NA \\ AUC = NA \\ PR = NA}\\
    Afib ischemic stroke & \makecell{\\Pre = 8.7 \\ AUC = 84.0 \\ PR = 48.5} & \makecell{\\Pre = 10.2 \\ AUC = 78.9 \\ PR = 41.2} & \makecell{\\Pre = 10.4 \\ AUC = 70.7 \\ PR = 39.1} & \makecell{\\Pre = 16.6 \\ AUC = 77.1 \\ PR = 50.5} & \makecell{\\Pre = 15.8 \\ AUC =68.9 \\ PR = 36.6} & \makecell{\\Pre = 10.8 \\ AUC = 76.8 \\ PR = 38.5} & \makecell{\\Pre = 9.7 \\AUC = 67.2 \\PR = 27.2}\\
    CAD CABG & \makecell{\\Pre = 7.1 \\ AUC = 88.4 \\ PR = 55.9} & \makecell{\\Pre = 4.1 \\ AUC = 81.5 \\ PR = 25.2} & \makecell{\\Pre = 4.4 \\ AUC = 52.9 \\ PR = 4.3} & \makecell{\\Pre = 7.2 \\ AUC = 84.7 \\ PR = 31.3} & \makecell{\\Pre = 4.9 \\ AUC = 73.5 \\ PR = 24.3} & \makecell{\\Pre = 4.0 \\ AUC = 79.0 \\ PR = 24.1
} & \makecell{\\Pre = 3.5 \\AUC = 81.5 \\PR = 44.4}\\
\hline
  \end{tabular}
  \captionsetup{width=.90\textwidth}
  \caption{Logistic regression model performance across different datasets. In each cell, three numbers were reported including the prevalence of the positive cases, ROC-AUC, and PR-AUC}
  \label{tab:prediction_model_results}
\end{table*}
\endgroup

\subsection{Privacy Evaluations}
We adopted the privacy evaluation framework outlined by Yan \textit{et al.} \cite{Yan_Brad_2022}, focusing on quantifying the risk of privacy breaches through various types of attacks, including membership inference, attribute inference, meaningful identity disclosure, and nearest neighborhood adversary attacks (NNAA). We detailed each metric in the subsequent sections and summarized the results in Table \ref{tab:privacy_metrics_results}, which include a comparison to the privacy risk quantification calculated by Yan \textit{et al.} using the University of Washington (UW) dataset. Our results are comparable to those from all models tested by Yan \textit{et al.}. Furthermore, according to Yan \textit{et al.}, a risk score below $0.333$ for all defined metrics indicate a low risk to privacy \cite{Yan_Brad_2022}.  Note that this evaluation framework requires a separate evaluation dataset not used in training. To ensure a fair evaluation, we held out $11,066$ patients ($0.5\%$ of the training set) to create an evaluation set. Subsequently, we re-trained the model using the remaining patients, following the procedure detailed in Section \ref{data_processing}. We then generated 1 million synthetic patients using a top\_p$=95\%$ specifically for evaluating the privacy metrics. 

\begingroup
\setlength{\tabcolsep}{4pt} 
\footnotesize
\begin{table*}[hb]
  \centering
  \begin{tabular}{c|c|cccccc}
  \toprule 
  \textit{Privacy Metric}  & \textit{CEHR-GPT} & \textit{medGAN} & \textit{medBGAN} & \textit{EMR-WGAN} & \textit{WGAN} & \textit{DPGAN} \\
  \midrule
Attribute Inference & 0.027 & 0.0078 & 0.0117 & 0.0680 & 0.0042 & 0.0136 \\
    Membership Inference & 0.1266 & 0.1506 & 0.1828 & 0.2966 & 
    0.1758 &
    0.0000 \\
    Meaningful Identity Disclosure &  0.002 & 0.0021 & 
    0.0027 &
    0.00361 &  0.0034 & 0.0004 \\
    NNAA Risk &  -0.0011 & 0.0008 &  0.0004 & 0.0198 &  0.0085 & 0.0017\\
\hline
  \end{tabular}
  \captionsetup{width=.90\textwidth}
  \caption{Privacy metrics computed on the synthetic data generated by CEHR-GPT. In addition, we include the benchmark metrics for the UW dataset from Yan, C. \textit{et al.}. \cite{Yan_Brad_2022}}
  \label{tab:privacy_metrics_results}
\end{table*}
\endgroup

\subsubsection{Membership Inference Attack}
In a membership inference attack, the objective is to determine whether a real patient record was utilized in the training of a generative model using the synthetic data \cite{Yan_Brad_2022, Theodorou2023}. Such attacks can occur if the model overfits to the training data, causing the synthetic data to resemble the training dataset more closely than the non-training dataset.

The attack dataset was constructed as follows: 1) We generated negative examples by randomly selecting 10,000 samples from the holdout set and labeling them as negative; 2) We created positive examples by randomly selecting 10,000 patients from the training set and labeling them as positive. We then matched the patients in the attack dataset to the training dataset by calculating Euclidean distance between the patients in attack dataset and training dataset. Following the guidelines suggested in \cite{Yan_Brad_2022}, we used a threshold distance of 5, below which patients were classified as positive (i.e., their data is present in the training dataset) and above which as negative (i.e., their data is not present in the training dataset). This procedure was repeated 100 times to ensure robustness. We evaluated the method by calculating recall, precision, and the F1 score based on the ground truth labels and the inferred predictions. The F1 score was then used to assess the risk associated with the data.  The full results can be found in Supplementary Materials \ref{mem_inf_supp}.

\subsubsection{Attribute Inference Attack}
In the attribute inference attack, we assumed the attacker has access to a group of target patients along with their demographic data, including age, gender, race, the year of their initial clinical visit, and prevalent clinical conditions like hypertension, abdominal pain, chest pain, etc. With these features, the attacker identifies the patient in the synthetic dataset with the closest attribute resemblance to the target individual. Then the attacker aims to use the sensitive attributes of the matched patient from the synthetic dataset to infer the corresponding sensitive attributes of the target patient.

To evaluate the attribute inference risk, we randomly sampled 150,000 patients from the real patient dataset as a target group. The search group comprised 1 million generated synthetic patients. To find the most similar patients from the search group, we created a set of common attributes including the demographic data along with the top 1\% of the most prevalent condition concepts, represented as one-hot encoded features. Then a k-nearest neighbors (KNN with k = 1) algorithm was applied to each target patient and a synthetic patient with the smallest Euclidean distance was found. Finally, we extracted the sensitive attributes (condition concepts not in the top 1\% tier) from matched target and synthetic patients. F1 scores were computed for the sensitive attributes of each matched patient pair and aggregated across all matched patients.

A baseline analysis was conducted by comparing real patients to other real patients. A result lower than the baseline suggests that the likelihood of finding synthetic data similar to a real patient is lower than finding a real patient similar to a real patient who share sensitive attributes. This implies a lower attribute inference risk which could be acceptable. Our training set was randomly divided into two halves, with 1 million real patients assigned as the target group, and the remaining 1 million as the search group. The matching process was consistently applied and an aggregated F1 score was computed. The full results can be found in Supplementary Materials \ref{att_inf_supp}.

\subsubsection{Meaningful Identity Disclosure Attack}
The meaningful identity disclosure risk measures the likelihood of identifying a synthetic record by linking it to patient records in an identified population dataset that contains the training data. This linkage could potentially reveal sensitive information about individuals in the population. Adopted from the metrics developed by \cite{el2020evaluating}, we can calculate the metrics using the following equation:
\small
\[
    max\Big(\frac{1}{N}\sum^n_{s=1}(\frac{1}{f_s}\times\frac{1+\lambda_s}{2}\times I_s \times R_s), \frac{1}{n}\sum^n_{s=1}(\frac{1}{F_s}\times\frac{1+\lambda_s}{2}\times I_s \times R_s)\Big)
\]
\normalsize
where $N$ denotes the size of the population cohort, $n$ denotes the size of the real sample, $s$ denotes the record index in the real sample, $f_s$ represents the number of records matched with record $s$ in the real sample and $F_s$ represents the number of records matched with record $s$ in the population cohort. $\lambda_s$ is the upper bound of the adjustment parameter for incorrect matches. Following the same configuration in \cite{el2020evaluating}, the midpoint between $\lambda_s$ and the maximum value of 1 is used to be more conservative. $I_s$ is the binary indicator if any real record is matched with the synthetic record. $R_s$ is another binary indicator if the adversary could use at least L\% sensitive attributes from matched synthetic records to learn something new about the target.

To assess the risk, we implemented the risk model applied by \cite{Yan_Brad_2022} on the CUIMC patient population. We used 3.7M as the population cohort, 2.3M patients as the real sample, and 1M synthetic patients as the target cohort. The matching was performed using 10 QID (Quasi Identifiers) which consist of age, gender, and race demographic attributes and 7 most common diseases in CUIMC patients. Moreover, we established $\lambda_s$ at 0.23, setting it as the upper bound to gauge the maximum risk level. Since we have around 17K attributes in total across the condition, drug, and procedure domains, to be aligned with the threshold in \cite{Yan_Brad_2022}, we chose 0.1 for L, which resulted in approximately 17 sensitive attributes. We further assessed the risk scores based on the matching across two generalization levels of age group (e.g. age 32 vs age group 30-40) and any combinations of 7 common diseases, resulting in 256 runs. The full results can be found in Supplementary Materials \ref{reid_supp}.

\subsubsection{Nearest Neighbor Adversary Attack}
The Nearest Neighbor Adversary Attack (NNAA) risk is a measure that assesses model overfitting by comparing the closeness of training and synthetic data. If the distance between the training and synthetic data is less than that between the evaluation and synthetic data, it indicates potential overfitting. This concept forms the basis for the metric developed by the researchers. Let's denote $S_T$, $S_E$, and $S_S$ to be random samples from the training, evaluation, and synthetic datasets. The risk score is defined as the following,
\[
    AA_{ES} - AA_{TS}
\]
where $AA_{ES}$ and $AA_{TS}$ are defined below,
\small 
\[
    AA_{ES} = \frac{1}{2} \Big( \frac{1}{n} \sum_{i=1}^n \mathbbm{1} \big(d_{ES}(I) > d_{EE}(i) \big) + \frac{1}{n} \sum_{i=1}^n \mathbbm{1} \big(d_{SE}(i) > d_{SS}(i) \big) \Big)
\]
\[
    AA_{TS} = \frac{1}{2} \Big( \frac{1}{n} \sum_{i=1}^n \mathbbm{1} \big(d_{TS}(i) > d_{TT}(i) \big) + \frac{1}{n} \sum_{i=1}^n \mathbbm{1}\big(d_{ST}(i) > d_{SS}(i) \big) \Big)
\]
\normalsize
Here, $d_{ES}(i)=\min_j||x_E^i - x_S^j||$ indicates the distance between the evaluation data point $x_E^i \in S_E$ and its nearest synthetic neighbor $x_S^j \in S_S$; $d_{EE}(i)=\min_{j, j \neq i}||x_E^i - x_E^j||$ represents the distance between the evaluation data point $x_E^i$ and its nearest neighbor within the same evaluation set $S_E$, excluding itself. Analogous definitions apply for $d_{SE}$, $d_{SS}$, $d_{TS}$, $d_{TT}$, $d_{ST}$, and $d_{SS}$. For our study, we randomly chose samples of 10,000 patients from the training, evaluation, and synthetic datasets to compute the metric as per the outlined formula. The full results can be found in Supplementary Materials \ref{nnaa_supp}.

\section{Discussion}
To the best of our knowledge, this is the first attempt to utilize GPT for generating time-series heterogeneous EHR data, while preserving patient privacy. Our main contribution lies in designing a novel patient representation that preserves a complete timeline of the patient's history, along with crucial visit details, thereby enabling GPT to create realistic patient sequences. Importantly, this representation facilitates seamless conversion back to the OMOP format, simplifying dissemination and analysis. This patient representation could serve as an effective messenger for transferring information across various standard data models. At present, our system is tailored to the OMOP format. However, it is designed with adaptability in mind, enabling us to seamlessly integrate new encoder/decoder pairs. This flexibility would facilitate the conversion of patient sequences to other widely-used data models, such as i2b2\cite{murphy2010serving}.

The study undertakes a three-tiered evaluation approach, systematically comparing synthetic and real datasets based on their marginal (column-wise distribution), conditional (co-occurrence relationship), and joint distributions (predictive performance). Concurrently, as the evaluation progresses through these three levels, there is a corresponding escalation in the complexity and challenge of the tasks involved. The outcomes of the KL divergence analysis revealed a nuanced relationship between the top k/p sampling strategies and the performance across the evaluation levels. Specifically, an increase in top k/p values enhanced performance in level 1 concept prevalence. However, an excessively high or low top k/p value adversely affects both level 2 co-occurrence metrics and level 3 machine learning predictions. This pattern suggests that including more tokens in the predictive distribution introduces greater uncertainty and a wider array of potential patient trajectory variations in the data generation process, thereby escalating the difficulty of achieving comparable performance outcomes.

The sampling strategies of top\_p=95\% and top\_k=300 seem to be most effective for generating the synthetic data. For instance, the synthetic data created with top\_p=95\% demonstrates the second smallest divergence in both dimension-wise distribution and co-occurrence relationship. Simultaneously, the corresponding synthetic cohorts successfully replicated the performance metrics in all predictive tasks, with the exception of CAD CABG. Finally, the significance of this patient representation transcends synthetic data generation; we believe it has the potential to establish the groundwork for integrating time into patient representations across diverse EHR-based deep-learning models.

\subsection{Loss of Temporal Information}
The reason that CEHR-GPT replicated the performance metrics of the prediction tasks can be attributed to the use of time tokens in its underlying patient representation. The majority of the prediction problems are phrased as “For a group of target patients who share similar characteristics, who would experience a particular medical event in one year from the index event?” in EHR research, therefore maintaining a complete patient timeline is crucial for time-sensitive cohort constructions \cite{ohdsi2019book}.

We claimed that the proposed patient representation had almost zero loss of temporal information, although this makes intuitive sense, there does not exist a formal metric to quantify this. To bridge this gap, we conceived a new metric named loss of temporal information (LOTI) to estimate the shrinkage of the patient timeline due to the use of the patient representation in an EHR dataset. Let's denote $T$ to be the time interval measured in days, ATT to be an artificial time token that represents a time interval (\(W_0\)), $F$ to be a function that maps $T$ to an ATT token (four days to \(W_0\)). In addition, let $G$ be the inverse function of $F$ that converts an ATT to $T$, moreover, we impose the constraint on G such that it takes the lower bound of ATT e.g., \(W_0 \implies 0\) days. Formally, we define LOTI as the expected difference between the original time interval $T$ and the reconstructed time interval $G(F(T))$ as the following,  
\[
  LOTI=E_{p(T)}\Big[T-G\big(F(T)\big)\Big]
\]
where $P(T)$ is the probability of $T$ observed in the training data defined as, 
\[
  P(T)=\frac{\textit{Freq of T}}{\Sigma\textsuperscript{T} \textit{Freq of T}}
\]
We computed LOTI for the patient representations utilized in CEHR-GPT and baseline models, shown in Figure \ref{patient_history_comparison}. As Table \ref{tab:loti} shows, CEHR-GPT has the least LOTI compared to the other patient representations while GPT-OUTPAT has a slightly higher time shrinkage because the inpatient duration was not retained. CEHR-BERT has a relatively large LOTI compared to the previous two representations due to the use of coarse ATT tokens. Finally, GPT-Vanilla has the most LOTI, which is equal to the expected length of the timeline in the source population due to the complete collapse of the timeline. 

\begingroup
\setlength{\tabcolsep}{6pt} 
\renewcommand{\arraystretch}{2} 
\begin{table*}[t]
  \centering
  \begin{tabular}{cccc}
  \toprule 
  \textit{Representation}  & \textit{Between visit ATT token} & \textit{Between inpatient span ATT token} & \textit{LOTI}\\
  \midrule
    CEHR-GPT & \makecell{\\Day token for $T \leq 1080$ \\ LT token for $T > 1080$ \\\\} & Day token & 7.739 \\
    GPT-OUTPAT & \makecell{Day token for $T \leq 1080$ \\ LT token for $T > 1080$ \\\\} & N/A & 7.962 \\
    CEHR-BERT &   \makecell{Day token for $T < 7$ \\ Week token for $7 \leq T < 30$ \\ Month token for $30 \leq T < 360$ \\ LT token $T \geq 360$}   &  N/A & 31.482 \\
    GPT-Vanilla & N/A & N/A & 111.164 \\
    \hline
  \end{tabular}
  \caption{Loss of Temporal Information for Different Patient Representations}
  \label{tab:loti}
\end{table*}
\endgroup

\subsection{Time Invariance and Sensitivity}
Across all synthetic datasets, the dimension-wise distribution (marginal distribution) and co-occurrence relationship were well preserved, regardless of the patient timeline's integrity within the models used. Even with a high LOTI, baseline models such as CEHR-BERT accurately mirrored both marginal distribution and co-occurrence relationship, yielding results similar to those from CEHR-GPT. This implies that these two measures may be largely time-invariant, unaffected by any shrinkage in the patient timeline. The rationale behind this is rooted in their construction methods, which either disregard or marginalize the temporal factor. Marginal distribution was constructed by counting the unique number of patients associated with the target concept, which was then normalized by a constant. The construction disregarded temporality, as the placement of a concept on the timeline was not a factor of consideration. The co-occurrence matrix was created using time initially but was marginalized after all co-occurring pairs were collected from the patient population. 

On the contrary, the predictive performance is extremely sensitive to any change made to the patient timeline as the cohort construction requires the integrity of the patient timeline. For example, HF readmission requires a 30-day prediction window from the index event (defined as the hospitalization episode with a heart failure diagnosis). Any shrinkage to the patient timeline will disrupt the construction of this cohort. The synthetic HF readmission cohort produced by CEHR-BERT showed a readmission rate of 100\% due to the shrinkage of the timeline in this patient representation; whereas, the actual expected rate of readmission should be approximately 25\%. Compared to CEHR-GPT, GPT-OUTPAT encoded time intervals between visits but did not preserve the duration of inpatient visits, therefore having a slightly higher LOTI. As a consequence, it showed reasonable performance in the cohorts (Hospitalization and Afib ischemic stroke), which had a 360-day prediction window and were thus less impacted by timeline shrinkage. However, in the case of HF Readmission (where a 100\% readmission rate was observed) and COPD Readmission (which identified no patients), GPT-OUTPAT was less successful. These cohorts used a short 30-day prediction window, making them highly sensitive to any distortions in the timeline, which likely led to synthetic patients not meeting the cohort selection criteria. Interestingly, the CAD CABG cohort presents a notable deviation from the general trend, where the CEHR-GPT dataset with top\_k=100 outperformed both the top\_p=95\% and top\_k=300 configurations. Additionally, the CEHR-BERT synthetic data accurately replicated the machine learning performance metrics as well. This indicates that the CAD CABG cohort was less affected by time shrinkage. 

Therefore, selecting the appropriate patient representation is pivotal in maintaining specific properties of the source data when generating synthetic data. The choice hinges on the intended application of synthetic data, ensuring that critical features and patterns inherent to the patient information are accurately reflected and retained.

\subsection{Time Sensitive Forecasting}
Because the patient representation encodes all the temporal information in the sequence, the trained GPT model could be used potentially for time-sensitive forecasting. We could prompt the trained GPT model with a patient history and estimate the time of the next visit via a Monte Carlo Sampling approach shown in the following equation, 
\[
    P(\delta_{t}|h) \approx \frac{\sum^n_{i=1}\mathbbm{1}\Big[M_{gpt}(h) = \delta_{t} \Big]}{n} 
\]
where $M_{gpt}$ denotes the GPT model, $h$ denotes a patient history, $\delta_t$ denotes any time interval, and $n$ represents the number of samples. Then we can use the expectation $E\big(\delta_{t}\big)$ as the predicted time interval. In addition, we can quantify the confidence by calculating the standard deviation e.g. $sd(\delta_{t})$. Similarly, we can predict the visit type ($v$)  using the same Monte Carlo approach.
\[
    P(v|E\big(\delta_{t}\big), h) \approx \frac{\sum^n_{i=1}\mathbbm{1}\Big[M_{gpt}\Big(E\big(\delta_{t}\big), h\Big) = v \Big]}{n} 
\]
Finally, we can predict the most likely medical event ($c$) given the predicted visit type $v$,
\[
    P(c|v, E\big(\delta_{t}\big), h) \approx \frac{\sum^n_{i=1}\mathbbm{1}\Big[M_{gpt}\Big(v, E\big(\delta_{t}\big), h\Big) = c \Big]}{n} 
\]
This approach goes beyond conventional prediction methods by not only forecasting future medical events but also determining the timing of the next visit and the specific medical events associated with that visit type. This predictive model could provide a more detailed and actionable timeline for patient care.

\subsection{Limitations}
While synthetic datasets demonstrated a high degree of similarity to source data, they are subject to certain known constraints. Firstly, a selection bias was present in the training data due to the constraints on sequence length, ranging from 20 to 512. This limitation resulted in the partial inclusion of patients with chronic conditions, which typically require a longer context window for accurate representation. While extending the context window of the model could potentially address this issue, it may introduce unforeseen effects. Finding the optimal configuration to accommodate a broader context window would require comprehensive experiments. 

Secondly, identifying an optimal sampling strategy for generating synthetic data remains a challenge due to the presence of numerous hyperparameters such as temperature, top\_p, and top\_k. These parameters, when used in conjunction, could yield a wide array of configurations. While the top\_p=95\% strategy showed the least divergence from the source data, it was unable to accurately replicate performance metrics for CAD CABG. As an interim solution, we may publish multiple versions of synthetic data, along with their corresponding performance metrics. This approach would allow researchers to select the most suitable dataset for their specific use case. 

Thirdly, the GPT model showed a propensity to over-represent prevalent concepts, skewing towards those with higher frequencies in the dataset. An illustration of this is seen in the synthetic data, where 78\% of patients had at least one outpatient visit, in contrast to the actual data where this figure was 73\%. This discrepancy indicates a bias in the model towards more common occurrences. In addition, the co-occurrence analysis also demonstrated that the synthetic reconstruction faithfully represents the frequent concept pairs in the original data but may be less effective at recovering the finer associations between rare concepts as shown in Figure \ref{Diabetes_network}. To address the over-representation of prevalent concepts by the GPT model, future work will look into implementing regularization techniques. One promising approach is adaptive regularization, which can be outlined in several steps: 1) Model Training: Begin by training the GPT model for a predetermined number of steps; 2) Sequence Generation and Analysis: Generate a sample of patient sequences from the trained model and calculate the marginal distribution of the concepts within these sequences; 3) Distribution Comparison and Adjustment Score Calculation: Compare the model-generated distribution to the empirical distribution derived from the actual data. From this comparison, calculate an adjustment score for each concept; 4) Logit Adjustment: Modify the logits for each concept in the model according to the calculated adjustment scores. By implementing this procedure, the influence of each concept on the model’s learning process would be adaptively modified during back-propagation, allowing for an update in the model parameters that takes into account the disparity between the generated and actual data distributions. This should help in reducing the bias towards over-represented concepts.

Furthermore, improving patient representation is an area for further development. Currently, the model's representation is limited to daily intervals and does not capture more precise measurements like hours or minutes. This limitation is particularly relevant for intensive care unit (ICU) data where time-sensitive decisions are critical. Furthermore, the current model framework assigns the first visit of every synthetic patient to the start of the year, as it only includes a year token to denote the commencement of patient history. To refine the accuracy of patient history commencement, integrating a month token alongside the year token is being considered to properly represent seasonality. This would provide a more accurate reconstruction of the starting point for a patient's first visit in the generated data. 

Lastly, there is a necessity to incorporate the death event within patient sequences. Including this event would allow the synthetic data to more accurately represent mortality, enhancing its utility for predictions related to patient outcomes and lifespan. These enhancements aim to create a more precise and clinically relevant synthetic dataset that better mirrors the complexities of real-world patient trajectories.

\section{Conclusion}
To our knowledge, this is the first attempt to utilize GPT for generating time-series EHR data. Our main contribution lies in the design of a patient representation that captures temporal dependencies among token types, enabling GPT to generate realistic patient sequences. Moreover, this representation allows for easy conversion back to the OMOP format. Comprehensive evaluations showed that the synthetic data effectively captures the intricate patterns present in EHR data.

\bibliography{main}
\end{multicols}

\newpage
\section{Supplementary Materials}
\subsection{Patient Representation comparison}
\begin{Figure}
    \centering
    \includegraphics[width=0.8\linewidth]{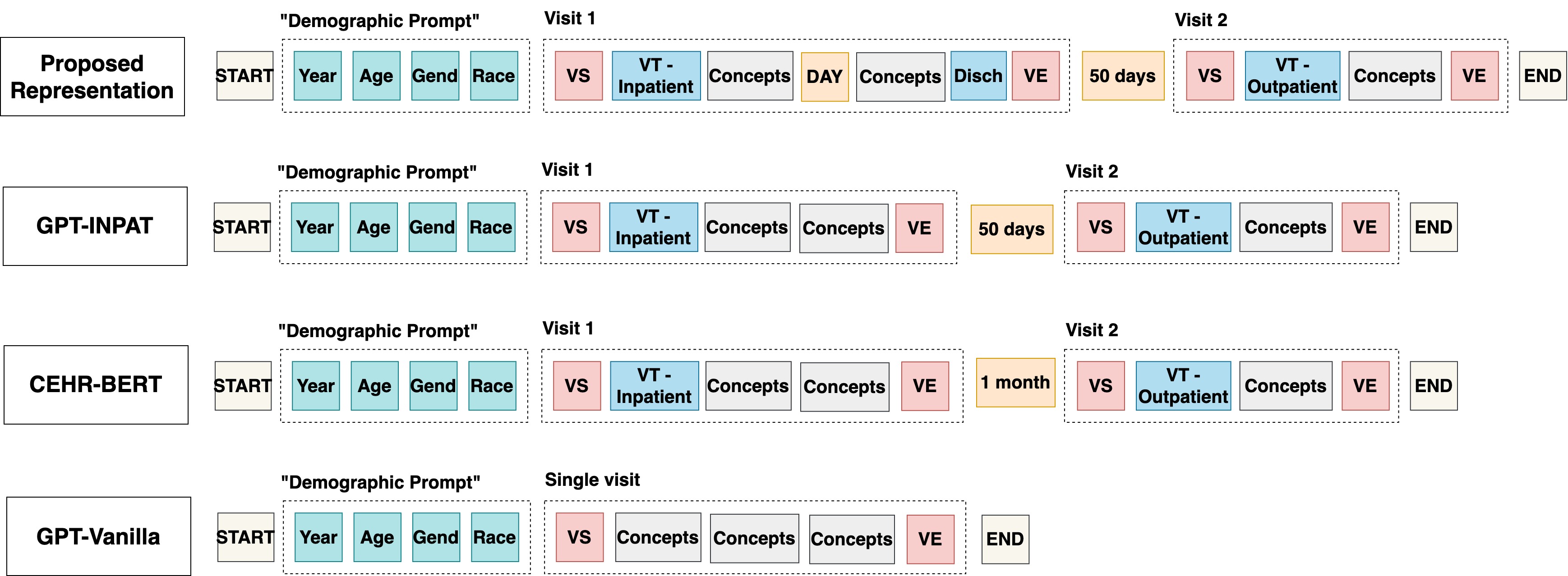}
    \captionof{figure}{The comparison of the patient representations. }
    \label{patient_history_comparison}
\end{Figure}

\newpage
\subsection{Synthetic cohort comparison by patient representation}
\begingroup
\setlength{\tabcolsep}{4pt} 
\footnotesize
\begin{table*}[tbh]
  \centering
  \begin{tabular}{c|c|cc}
  \toprule 
  \textit{Cohort}  & \textit{Real} & \textit{GPT-OUTPAT} & \textit{CEHR-BERT}\\
  \midrule
HF readmission & \makecell{\\Pre = 25.7\\AUC = 65.7\\PR = 39.3} & \makecell{\\Pre =100.0 \\ AUC = NA \\ PR = NA} & \makecell{\\Pre =100.0 \\ AUC = NA \\ PR = NA}\\
    Hospitalization & \makecell{\\Pre = 5.6 \\ AUC = 75.3 \\ PR = 19.5}  & \makecell{\\Pre = 5.3 \\ AUC = 70.2 \\ PR = 14.3} & \makecell{\\Pre = 0.3 \\ AUC = 76.0 \\ PR = 1.0}\\
    COPD readmission &  \makecell{\\Pre = 34.5 \\ AUC = 74.2 \\ PR = 83.8} & \makecell{\\Pre = NA \\ AUC = NA \\ PR = NA}  &  \makecell{\\Pre = NA \\ AUC = NA \\ PR = NA}\\
    Afib ischemic stroke & \makecell{\\Pre = 8.7 \\ AUC = 84.0 \\ PR = 48.5} & \makecell{\\Pre = 9.7 \\AUC = 67.2 \\PR = 27.2} & \makecell{\\Pre = 19.5 \\AUC = 72.4 \\PR = 60.0} \\
    CAD CABG & \makecell{\\Pre = 7.1 \\ AUC = 88.4 \\ PR = 55.9} & \makecell{\\Pre = 3.5 \\AUC = 81.5 \\PR = 44.4} & \makecell{\\Pre = 5.4 \\AUC = 77.2 \\PR = 44.1}\\
\hline
  \end{tabular}
  \captionsetup{width=.60\textwidth}
  \caption{Logistic regression (LR) model performance across baseline synthetic datasets. In each cell, three numbers were reported including the prevalence of the positive cases, ROC-AUC, and PR-AUC. If the prevalence=NA, this indicates that $0$ patients were identified in the cohort. AUC=NA and PR=NA indicate the LR model could not be run successfully due to either 100\% or 0\% prevalence. } 
  \label{tab:supple_prediction_model_results}
\end{table*}
\endgroup

\subsection{Visit Prevalence analyses}
\begin{Figure}
    \centering
    \includegraphics[width=0.8\linewidth]{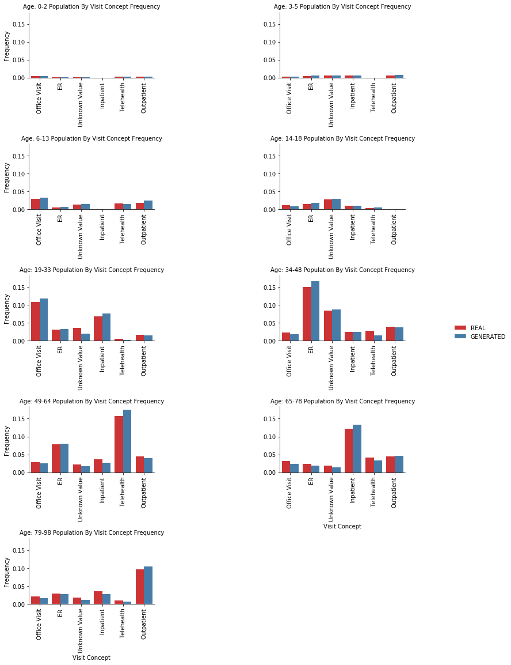}
    \captionof{figure}{The visit prevalence stratified by age group}
    \label{visit_age_breakdown}
\end{Figure}

\begin{Figure}
    \centering
    \includegraphics[width=0.8\linewidth]{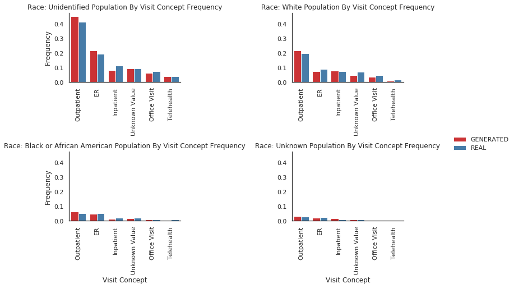}
    \captionof{figure}{The visit prevalence stratified by race}
    \label{visit_race_breakdown}
\end{Figure}

\begin{Figure}
    \centering
    \includegraphics[width=0.8\linewidth]{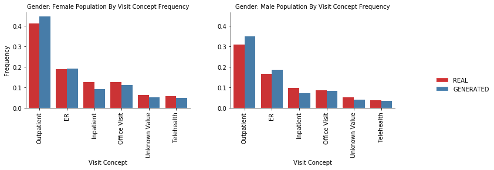}
    \captionof{figure}{The visit prevalence stratified by gender}
    \label{visit_gender_breakdown}
\end{Figure}

\newpage

\subsection{Privacy Evaluation Metrics}
\subsubsection{Membership Inference} \label{mem_inf_supp}
Table \ref{tab:membership_inference_results} shows that the accuracy of both attacks is slightly less than 50\%, indicating that the performance of such attacks is worse than a random guess.

\begin{center}
\renewcommand{\arraystretch}{1.2} 
  \begin{tabular}{ccccc}
  \toprule 
  \textit{}  & 
  \textit{Accuracy} & \textit{Recall} & \textit{Precision} & \textit{F1}\\
  \midrule
  Data Attack & $0.4997 \pm 0.0008$ & $0.0725 \pm 0.0026$ &  $0.4996 \pm 0.0055$ & $0.1266 \pm 0.0040$\\
    \hline
  \end{tabular}
  \captionof{table}{Membership Inference Attack metrics}
  \label{tab:membership_inference_results}
\end{center}

\subsubsection{Attribute Inference} \label{att_inf_supp}
Table \ref{tab:attribute_inference_results} shows that the F1 score of synthetic vs real is less than real vs real scenario. But it's still higher than other models that didn't capture the real patterns as effectively. \cite{Theodorou2023}

\begin{center}
\renewcommand{\arraystretch}{1.2} 
  \begin{tabular}{ccccc}
  \toprule 
  \textit{}  & 
  \textit{Recall} & \textit{Precision} & \textit{F1}\\
  \midrule
  Synthetic vs Real & 0.0350 & 0.0343 & 0.0271\\
  Real vs Real & 0.0612 & 0.0468 &  0.0421\\
    \hline
  \end{tabular}
  \captionof{table}{Attribute Inference Attack metrics}
  \label{tab:attribute_inference_results}
\end{center}

\subsubsection{Meaningful Identity Risk} \label{reid_supp}
Table \ref{tab:reid} displays the maximum risk scores for two terms in the equation, calculated across all 256 combinations.  The ultimate risk score, derived from the highest of these two components, is 0.002053. This risk score is significantly below the threshold of 0.09 proposed by the creator of this metric El Emam et al \cite{el2020evaluating} to determine whether the synthetic data violates privacy. 

\begin{center}
\renewcommand{\arraystretch}{1.2} 
  \begin{tabular}{cc}
  \toprule 
  \textit{First Term}  & 
  \textit{Second Term} \\
  \midrule
    0.001353 & 0.002053 \\
    \hline
  \end{tabular}
  \captionof{table}{Meaningful Identity Disclosure Attack}
  \label{tab:reid}
\end{center}

\subsubsection{Nearest Neighbor Adversary Attack} \label{nnaa_supp}
This procedure was conducted five times to obtain an average distance difference $-0.0011$ as shown in Table \ref{tab:nnaa}. Following the guidelines in \cite{Yan_Brad_2022}, an NNAA risk below 0.03 is deemed minimal, validating the privacy adequacy of our synthetic data.

\begin{center}
\renewcommand{\arraystretch}{1.2} 
  \begin{tabular}{cc}
  \toprule 
  \textit{}  & 
  \textit{NNAA} \\
  \midrule
   Run 1 & -0.0031 \\
   Run 2 & -0.0012 \\
   Run 3 & -0.0045 \\
   Run 4 & 0.0006 \\
   Run 5 & 0.0024 \\
   \hline
  \end{tabular}
  \captionof{table}{Nearest Neighbor Adversary Attack}
  \label{tab:nnaa}
\end{center}

\end{document}